\documentclass[letterpaper, 10 pt, conference]{ieeeconf}  % Comment this line out if you need a4paper

\IEEEoverridecommandlockouts                              % This command is only needed if 
\usepackage{graphicx} % for pdf, bitmapped graphics files
\usepackage{amsmath} % assumes amsmath package installed
\usepackage{amssymb}  % assumes amsmath package installed

\usepackage[font=footnotesize]{caption}
\usepackage{nameref}
\usepackage{siunitx}
\usepackage{subcaption}
\usepackage{booktabs}
\usepackage{hyperref}
\usepackage{pgfplots}
\usepackage{tikzscale}
\pgfplotsset{compat=newest}
\newlength\figureheight
\newlength\figurewidth 

\usepackage[noadjust]{cite}

\title{\LARGE \bf
	Dynamic Occupancy Grid Prediction for Urban Autonomous Driving:\\
	A Deep Learning Approach with Fully Automatic Labeling
}

\author{Stefan Hoermann$^{1}$, Martin Bach$^{1}$ and Klaus Dietmayer$^{1}$ 
\thanks{The authors are with:
\newline
$^{1}$ Institute of Measurement, Control, and Microtechnology, Ulm University, Germany
	{\tt\small \{firstname.lastname\}@uni-ulm.de} %
}%
}

\begin{document}

\bstctlcite{IEEEexample:BSTcontrol} 

\def\cred{\textcolor{red}}
\def\cblue{\textcolor{blue}}
\def\cgreen{\textcolor{green}}
\def\etal{et\:al.\ }
\def\DOGMA{DOGMa}

% % Symbols definition
% particle
\def\particle{\rho}

% grid cell
\def\gridcell{c}

% Channel
\def\OUTchannel{\mathcal{T}}
\def\GMchannels{\Omega}
\def\GMwidth{W}
\def\GMheight{H}

\def\targettime{k}

\def\Occupied{\mathrm{O}}
\def\MassFree{M_\mathrm{F}}
\def\MassOcc{M_\mathrm{O}}
\def\East{\mathrm{E}}
\def\North{\mathrm{N}}

\def\static{\mathrm{s}}
\def\dynamic{\mathrm{d}}

\def\rocthreshold{\gamma}

\maketitle
\thispagestyle{empty}
\pagestyle{empty}

%%%%%%%%%%%%%%%%%%%%%%%%%%%%%%%%%%%%%%%%%%%%%%%%%%%%%%%%%%%%%%%%%%%%%%%%%%%%%%%%
\begin{abstract}
Long-term situation prediction plays a crucial role in the development of intelligent vehicles.
A major challenge still to overcome is the prediction of complex downtown scenarios with multiple road users, e.g., pedestrians, bikes, and motor vehicles, interacting with each other.
This contribution tackles this challenge by combining a Bayesian filtering technique for environment representation, and machine learning as long-term predictor.
More specifically, a dynamic occupancy grid map is utilized as input to a deep convolutional neural network.
This yields the advantage of using spatially distributed velocity estimates from a single time step for prediction, rather than a raw data sequence, alleviating common problems dealing with input time series of multiple sensors.
Furthermore, convolutional neural networks have the inherent characteristic of using context information, enabling the implicit modeling of road user interaction.
Pixel-wise balancing is applied in the loss function counteracting the extreme imbalance between static and dynamic cells.
One of the major advantages is the unsupervised learning character due to fully automatic label generation.
The presented algorithm is trained and evaluated on multiple hours of recorded sensor data and compared to Monte-Carlo simulation. 
\end{abstract}

%%%%%%%%%%%%%%%%%%%%%%%%%%%%%%%%%%%%%%%%%%%%%%%%%%%%%%%%%%%%%%%%%%%%%%%%%%%%%%%%

% *** Section 1 - Introduction ***
\section{Introduction}
\label{sec:Introduction}

Research of the past decades gained great success in the field of autonomous driving. 
Comparing automated vehicles to human drivers, humans can compensate their comparably long reaction time with the ability to estimate a future scene evolution.
This also enables strategic and forward-looking behavior, hard to compensate with the fast reaction and precise perception of machines.
Long-term situation prediction for automated driving is therefore a major challenge still to overcome.
We tackle a highly complex urban environment where cars, trucks, bikes and pedestrians share the road.
Road users react individually, but have type-specific motion constraints.
Most important, humans, walking or driving, do not act independently in such scenarios, but rather interact and account for possible behavior of others.
Consequently, an individual decision and path may have consequences to the future behavior of someone or everyone else.
Such traffic scenarios might, hence, evolve into many possible future scene constellations, motivating the need for a probabilistic representation of the prediction result.
In short, long-term situation prediction for urban autonomous driving should be able to incorporate interactions of all known traffic participants and, due to the uncertain nature of the problem, handle the variance of possible outcomes.

% % % % % % % % % % % % % % % % % % % % % % % %
% GRID MAP FIGURE
% % % % % % % % % % % % % % % % % % % % % % % %
\begin{figure}
\centering\includegraphics[angle=-90,width=8cm]{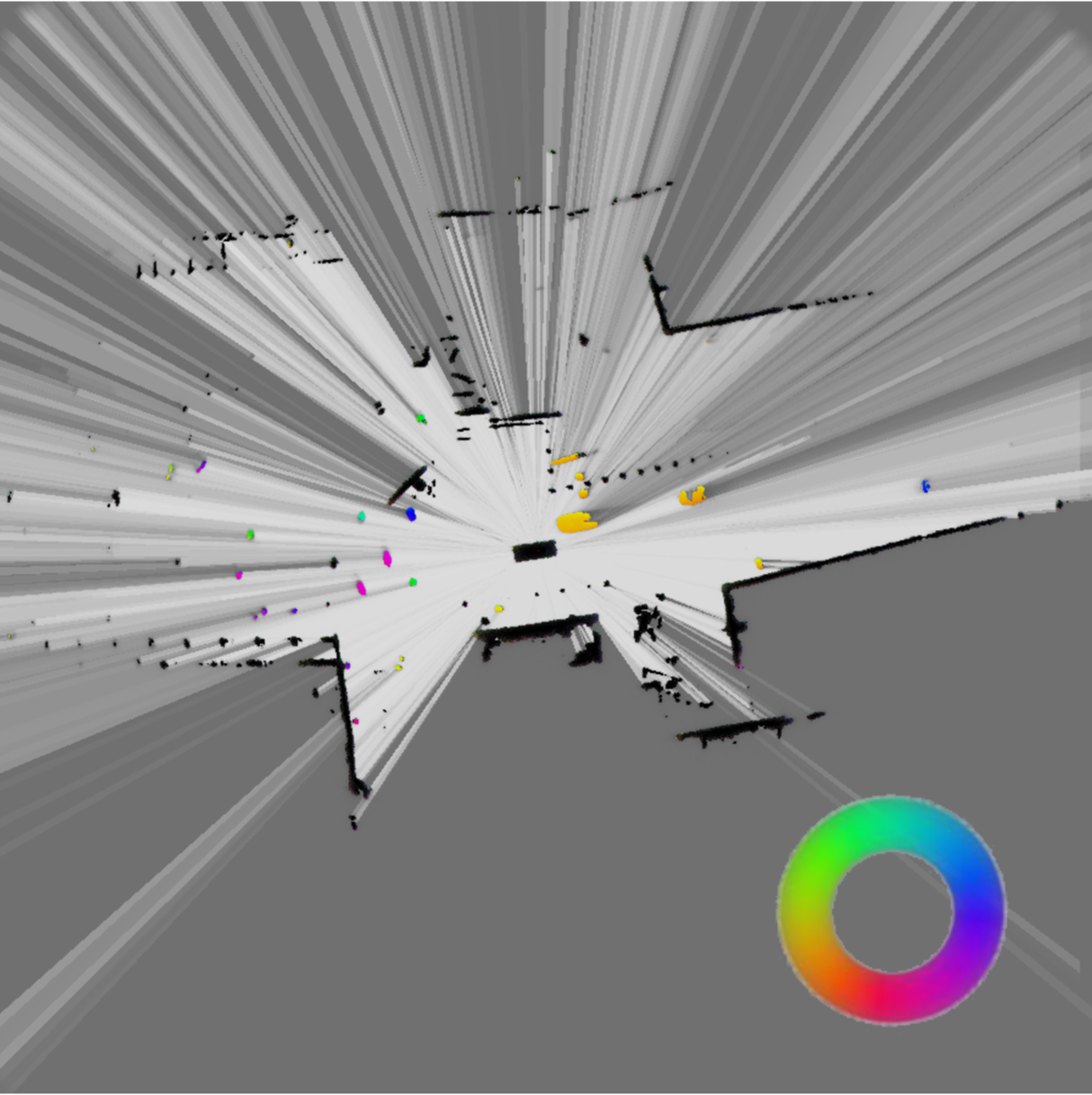}
\caption{Visualization of a Dynamic Grid Map result. The colored cells indicate the velocity orientation according to the colored circle. Grayscale cells indicate the occupancy probability where black indicates occupied space. The occupied rectangle in the center of the image is the ego vehicle.}
\label{fig:dynamci_grid_map}
\end{figure}
Models of \emph{interactions} and \emph{motion dynamics} are common tools to improve prediction.
Literature covers long-term prediction using engineered and learned models.
Manually designed dependencies in road user prediction, e.g., using a digital map or the Intelligent Driver Model \cite{kesting2008car_following_models}, usually underlie quite limiting assumptions, e.g., a lane following \cite{petrich2013mapbasedprediction} or car following \cite{hoermann2017} scenario.
In addition, hand crafted models considering context and object relations tend to be high dimensional \cite{kesting2008car_following_models, odhams2004speedincurves}, which motivates machine learning as an alternative.
Concordant to the great success of machine learning for classification, methods were proposed to predict a future maneuver with regard to discrete maneuver classes, e.g., \textit{stop}, \textit{go-straight} or \textit{turn} \cite{Graf_Intersection_2014, Kuhnt2016}.
However, predicting a trajectory in terms of a time series differs from predicting a maneuver class in terms of generalization.
To gain a spatio-temporal distribution at the algorithm output, it is, however, a matter of data representation. 
Wiest \cite{wiest2017PHD}, e.g., predicted parameters of Chebyshev polynomials considered as random variables in a Gaussian mixture model. 

In contrast to the algorithm output, considering time series of raw sensor data at the input, several technical problems arise despite data representation.
Varying sample frequency, asynchronous sensors, scaling and translation in time domain, or out of sequence input are examples which are challenging for machine learning \cite{Laengkvist201411}, but well-studied by the sensor fusion community \cite{barshalom2001OutOfSequence,aeberhard2012AsynchronousSensors}.
From a machine learning perspective, instead of learning temporal features from time series and facing the aforementioned problems, state variables of a single time stamp are related by first-order differential equations representing short-term dynamic features.
In addition, we assume, that a lack of long-term input sequences can be compensated by spatial context.
This assumption is inspired by research in stock market prediction \cite{agrawal2013stockpredictiontechniques} where it is known that using the past time series is not sufficient to predict the future stock trend \cite{tsai2010stockprediction}.
Moreover, it is not clear if they are helpful at all \cite{fama1965stockprediction_isguessing}.
However, prediction performance can be improved \cite{agrawal2013stockpredictiontechniques} by context information like twitter moods \cite{bollen2011twitter_predicts_stock} or online chats \cite{gruhl2005predictstockbyonlinechat}.

The spatial context can be seen in a dynamic occupancy grid map (\DOGMA), as illustrated in Fig. \ref{fig:dynamci_grid_map}.
The \DOGMA{} uses Bayesian filtering \cite{NUSS2016} to fuse a variety of sensors and represents the entire environment, static and dynamic, in a bird's eye view.
Each multichannel pixel, or cell, contains probabilistic occupancy and velocity estimates.
Moreover, the image-like data structure of \DOGMA s naturally suggests using convolutional neural networks (CNNs) to exploit contextual information and dependencies between different objects and the given infrastructure.

% % %
Therefore, in this work, the strengths of engineered and learning based methods are combined. 
A convolutional neural network (CNN) is used to model long-term motion, and a Bayesian estimate of the current dynamic environment is used at the input.
For training, no hand crafted labeling has to be performed, however, an algorithm separating rare dynamic from static grid cells was applied using future and past estimates.
First of all, this enables balancing between static and dynamic regions during training, but moreover a learned segmentation of the current environment outperforming engineered approaches.

The remaining paper is organized as follows.
In Section \ref{sec:gridmap}, we give an overview of dynamic occupancy grid mapping. 
Section \ref{sec:network_output_labels} explains fully automatic label generation for unsupervised learning.
\ref{sec:cnnArchitecture} gives a short overview of the employed CNN architecture.
In Section \ref{sec:loss_function}, a pixel balancing loss function is introduced, counteracting the high underrepresentation of dynamic cells.
Results are shown in Section \ref{sec:results} followed by conclusions in \ref{sec:conclusion}.

% *** Section 2 
\section{Filtered Dynamic Input}
\label{sec:gridmap}

A dynamic occupancy grid map (\DOGMA) \cite{NUSS2016}, fed by multiple sensors, can provide a bird's eye image like \SI{360}{\degree} representation of the environment, as illustrated in Fig. \ref{fig:dynamci_grid_map}. 
The \DOGMA\ data is provided in $\mathbb{R}^{|\GMchannels| \times \GMwidth \times \GMheight}$ with the spatial width $\GMwidth$ and height $\GMheight$ pointing in east and north direction, respectively.
The channels $\GMchannels = \left\{ \MassOcc, \MassFree, v_\East, v_\North, \sigma^2_{v_\East}, \sigma^2_{v_\North}, \sigma^2_{v_\East, v_\North} \right\}$ contain the Dempster-Shafer masses for free space $\MassFree \in \left[0,1\right]$ and occupancy $\MassOcc \in \left[0,1\right]$, the velocity pointing east $v_\East$ and north $v_\North$, as well as the velocity variances and covariance.
The occupancy probability is calculated by 
\begin{equation}
P_\Occupied = 0.5\cdot\MassOcc + 0.5\cdot(1-\MassFree)
\end{equation}
where a high $P_\Occupied$ refers to a dark pixel in Fig. \ref{fig:dynamci_grid_map}.
All channels $\GMchannels$ and extensions can be used to feed a neural net, although Fig. \ref{fig:dynamci_grid_map} only contains $3$ RGB-channels for illustration.

Feeding raw sensor data into the CNN is intentionally avoided, due to two main advantages:
1) \DOGMA\ cells provide a velocity estimate with covariance matrix for each cell resulting in a spatial occupancy and velocity distribution. For this, the algorithm uses advanced sensor fusion techniques regarding latency and variance in time domain which are hard to compensate by learning based approaches \cite{Laengkvist201411}.
2) The \DOGMA\ output format is independent of the sensor setup, e.g., fusing radar and laser or only lasers with different range, which can be advantageous when training a neural network with varying sensors.
In general, the algorithm exploits sensor dependent strengths using engineered models.
These models can, e.g., consider free space gained from a laser measurement, or ego motion compensated Doppler velocities from radars \cite{NUSS2016}.
If sensor specifications change, parameters can easily be adapted.

The \DOGMA\ is based on Sequential Monte Carlo filtering.
Grid cells have a static position and width, and the map border is shifted by the cell width to keep the moving ego vehicle within the center cell.
The filter algorithm runs in two modules:
In the first, sensor data is converted into classical occupancy grid maps \cite{Elfes1989, Thrun2005} using sensor specific measurement models.
These grids serve as a generic data interface to the second module, a particle filter estimating the spatial occupancy and velocity distribution \cite{NUSS2016}.
Each grid cell is updated independently and no interaction between cells is modeled.
Although, hand-crafted methods were proposed to extract modeled objects from grid cells \cite{steyer2017object}, in this work, no object generation is needed at the CNN input or for labeling.
Due to the implementation in the Dempster-Shafer domain \cite{dempster2008generalization}, a \DOGMA\ provides probabilistic information about occupancy, free space, and unobservable areas.
%

% *** Section 3 
\section{Automatic Output Label Generation}
\label{sec:network_output_labels}

Labeling is usually an extensive process and one of the major drawbacks in supervised learning.
In the nature of scene prediction, fortunately, the desired algorithm output can be observed at a later time.
Once automatic label extraction can be applied, it is comparably easy to generate training data.
However, about $99.75\%$ of our environment data is static, which might lead the learning algorithm to favor the trivial result of predicting the future to be equal to the current input sample.
For this reason, a segmentation of the underrepresented dynamic environment is required to apply balancing methods.
Methods to classify dynamic cells online at perception time are proposed by \cite{NUSS2016}. 
However, using the current cell velocity estimate leads to misclassification when static objects enter the field of view or at noisy regions, e.g., the borders of static objects.
Since the label extraction can be done offline, a better classification can be achieved using future and past occupancy course, and thus, e.g., distinguish between growing static parts and actually moving objects.

\begin{figure}[t]
\vspace{1.5mm}
\centering\includegraphics{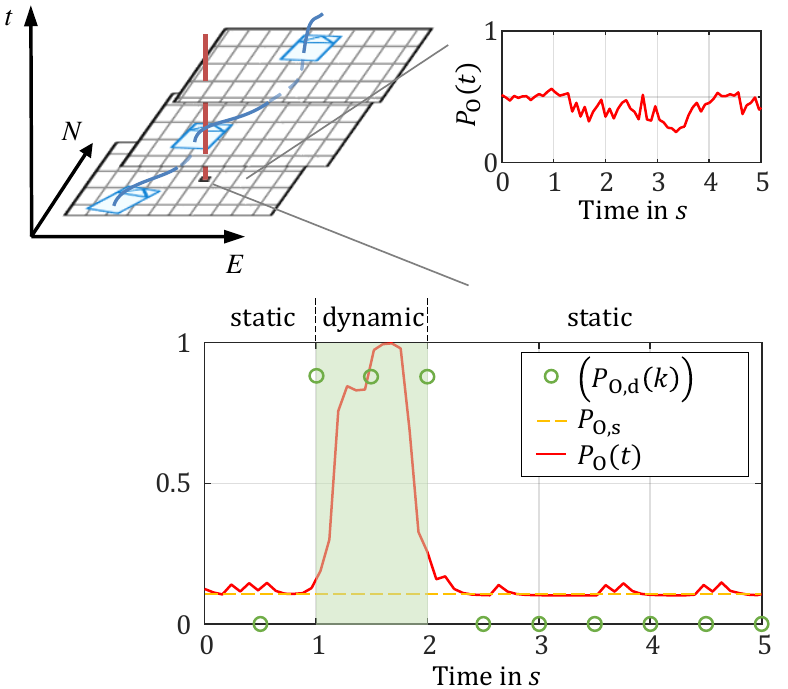}
\caption{Extracting the occupancy probability course of a single cell with fixed spatial position from a grid map with moving objects when the grid map origin is also moving.}
\label{fig:single_cell_P_o_verlauf}
\end{figure}

For each cell, a quasi time continuous $P_\Occupied(t)$ is extracted, as illustrated in Fig. \ref{fig:single_cell_P_o_verlauf}. 
The figure illustrates the extraction of $P_\Occupied(t)$ for grid cells at constant spatial coordinates $(E,N)$, while the ego vehicle is itself moving in a dynamic environment.
The desired prediction result is a sequence
\begin{equation}
\left(P_\Occupied(k)\right)= \left(P_\Occupied(1), P_\Occupied(2), P_\Occupied(3),...\right)
\end{equation}
where the time step sequence $k=(1,2,...)$ can be represented in corresponding output channels of the CNN. 
In addition, a segmentation into static and dynamic parts is desired.
Thus, one can write 
\begin{equation}
\label{eq:POCC_timeseries}
\begin{split}
\left(P_\Occupied(k)\right) &= P_{\Occupied,\static} +  \left(P_{\Occupied,\dynamic}(1),  P_{\Occupied,\dynamic}(2),P_{\Occupied,\dynamic}(3),... \right) \\
&= P_{\Occupied,\static} + \left(P_{\Occupied,\dynamic}(k)\right)\\
\end{split}
\end{equation}
with a constant occupancy probability $P_{\Occupied,\static}$ of the static environment and a sequence $\left(P_{\Occupied,\dynamic}(k)\right)$ describing the occupancy probability originated solely from dynamic elements.
The segmentation can be done automatically in time domain by dividing data cell-wise in static and dynamic time $t_\static$ and $t_\dynamic$:
When objects traverse a cell, the Bayesian filter in the \DOGMA{} converges smoothly to a maximum $P_\Occupied$ and back to the static level when the object leaves the cell (see Fig. \ref{fig:single_cell_P_o_verlauf}). 
A rise in $P_\Occupied(t)$ followed by a drop of the signal is used to detect the presence of a dynamic object, and thus the time in between defines the interval $t_\dynamic$.
This calculation is based on the second derivative followed by non-maximum suppression.
In the offline extracted sequence, the \DOGMA{} filter convergence delay is corrected by setting $\left(P_{\Occupied}(k)\right)$ to $\max\left(P_\Occupied(t_\dynamic)\right)$ for time steps within $t_\dynamic$.
The curve $P_\Occupied(t)$ can be quite noisy, especially in unknown regions ($P_\Occupied=0.5$) or at the spatial transition between static, free, and unknown. 
Therefore, the data in $\GMchannels$ is smoothed with a Gaussian filter in time domain.
Noise in spatial domain, e.g., a single dynamic cell, is smoothed accordingly.

Finally, the constant static $P_{\Occupied,\static}$ in (\ref{eq:POCC_timeseries}) is calculated by the median of $P_{\Occupied,\static}(t_s)$ with $t_s \in (t_0,t_0+T) \setminus t_d$, where $t_0$ denotes the \DOGMA{} input time and $T$ the prediction horizon.
The network output in $\mathbb{R}^{|\OUTchannel| \times \GMwidth \times \GMheight}$ provides the channels
$\OUTchannel = \left\{ P_{\Occupied,\static}, P_{\Occupied,\dynamic}(1), P_{\Occupied,\dynamic}(2), ..., P_{\Occupied,\dynamic}(|\OUTchannel|-1) \right\}$. % with $P_{\Occupied,\dynamic}(|\OUTchannel|-1) = P_{\Occupied,\dynamic}(T)-P_{\Occupied,\static}$.

% *** Section 4
\section{CNN Architecture}
\label{sec:cnnArchitecture}
A deep neural network architecture can exploit far distant relations within the \DOGMA\ input.
However, the output structure should also provide a spatial distribution.
The network architectures of \cite{noh2015deconvnet} (DeconvNet) and \cite{shelhamer2017FCNs} (Fully Convolutional Networks, FCN), provide upscaling layers following downscaling cascades.
These approaches aim at pixel precise image segmentation from a single input image.
The group around Shelhamer and Long \cite{shelhamer2017FCNs, long2015FCN} resizes down sampled feature maps using deconvolution operations proposed by Zeiler and Fergus \cite{zeiler2014FCNvis} for visualization purposes.
Upsampling is performed in a single layer while earlier layer results with a higher resolution are used to gain a fine output map.
In contrast to learnable kernels, the authors of \cite{noh2015deconvnet} and \cite{badrinarayanan2015segnet} use the unpooling method of \cite{zeiler2011adadeconv}.
While \cite{shelhamer2017FCNs} use a single upsampling step, \cite{noh2015deconvnet} perform upscaling stepwise, mirroring the downsampling stages and reusing the indices of max-pooling downscaling for upscaling by unpooling.
The resulting sparse feature maps are made dense by subsequent convolution in each unpooling layer.
In our approach, we follow the strategy of a single input \DOGMA\ processed by a downscaling cascade mirrored by upscaling stages, as proposed by \cite{noh2015deconvnet}.
However, we chose learnable deconvolution kernels instead of unpooling, combined with bypassed results from according downscaling stages.

Instead of pixel classification we aim at the prediction of occupied cells at future time steps.
Moreover, the network segments static and dynamic parts of the environment and delivers the occupancy of static grid cells in one channel and the occupancy of dynamic cells in the remaining channels.
%

% *** Section 5
\section{Spatial Balancing Loss}
\label{sec:loss_function}

Due to the segmentation of the labels in static and dynamic areas, balancing methods can be applied to weight rare ($0.25\%$) dynamic cells more compared to static cells.
In our specific case, the network is also trained to perform a segmentation of static and dynamic regions.
Therefore, the occupancy information of static cells is stored in the first channel, while only occupancy of rare dynamic cells is included in the remaining time channels.
The time channel labels of static cells are set to $0$.
For one sample, the loss
\begin{equation}
L = L_\static + L_\dynamic
\end{equation}
is the sum of a static channel loss $L_\static$ and the loss of dynamic channels $L_\dynamic$.
The linear least squares of the single static output channel in $\OUTchannel$ is used to calculate the loss of static occupancy estimation for each grid cell $\gridcell$ with
\begin{equation}
L_\static = \frac{1}{2} \lambda_\static \sum_{\gridcell=1}^{\GMwidth\times\GMheight}\left(P_{\Occupied,\static}(\gridcell) - \hat{P}_{\Occupied,\static}(\gridcell)\right)^2
\end{equation}
with $\lambda_\static$ denoting a static weighting factor. 
We write the labeled pixel value with $P_{\Occupied,.}$ and the network output $\hat{P}_{\Occupied,.}$.

For the dynamic channels we calculate a cell wise weighting factor 
\begin{equation}
\lambda_{\gridcell,\targettime} = \lvert 1+\lambda_{\targettime} P_{\Occupied,\dynamic}(\gridcell,\targettime)\rvert
\end{equation}
for each time channel $\targettime$ and grid cell $\gridcell$.
Since $P_{\Occupied, \dynamic}(\gridcell,\targettime) \in \left[0,1\right]$ with $P_{\Occupied, \dynamic}(\gridcell,\targettime) = 0$ at static cells, $\lambda_{\gridcell,\targettime} = 1$ for static and $\lambda_{\gridcell,\targettime} = 1+\lambda_{\targettime}$ for dynamic regions.
The factor $\lambda_{\targettime}$ is used to apply increasing weighting factors for the prediction time channels $\targettime$.
The dynamic loss is then calculated with 
\begin{equation}
\label{eq:dynamic_loss}
L_\dynamic = \frac{1}{2} \sum_{\targettime=1}^{|\OUTchannel|-1}  \sum_{\gridcell=1}^{\GMwidth\times\GMheight} \lambda_{\gridcell,\targettime} \left(P_{\Occupied,\dynamic}(\gridcell,\targettime) - \hat{P}_{\Occupied,\dynamic}(\gridcell,\targettime)\right)^2
\end{equation}
and contains an inherited discrimination between static and dynamic.
For back propagation, the partial derivative of (\ref{eq:dynamic_loss}) at a cell $\gridcell$ and time channel $\targettime$ is calculated with
\begin{equation*}
\label{eq:dynamic_loss_derivative}
\frac{\delta L_\dynamic}{\delta \hat{P}_{\Occupied,\dynamic}} = -\lvert 1+\lambda_\targettime P_{\Occupied,\dynamic}(\gridcell,\targettime)\rvert 
\left(P_{\Occupied,\dynamic}(\gridcell,\targettime) - \hat{P}_{\Occupied,\dynamic}(\gridcell,\targettime)\right) ~ .
\end{equation*}
%

% *** Section 6
\section{Dataset and Training}
\label{sec:dataset_and_training}

\begin{figure}
\vspace{1.7mm}
\centering\includegraphics[width=8cm]{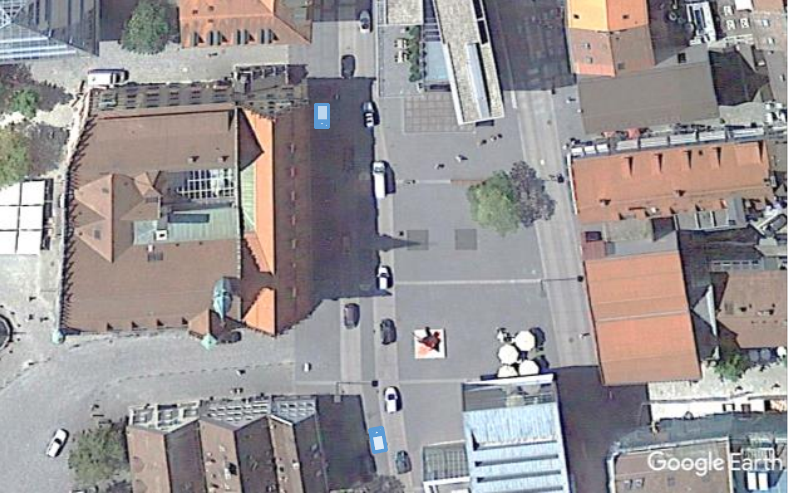}
\caption{Satellite image of the shared space scenario. At three different days, sequences were recorded from the two positions drawn in blue.}
\label{fig:Neue_Mitte_satellite}
\end{figure}

We recorded about \SI{2.5}{h} of an urban shared space scene with pedestrians, bikes and motor vehicles at three different days.
At every day, the recording vehicle was placed at two different positions, facing west and east, as illustrated in Fig. \ref{fig:Neue_Mitte_satellite}.
The update rate of the Dynamic Grid Map is about \SI{100}{ms} resulting in $77582$ samples in total.
We used $62800$ samples for training, $7590$ for testing and $7192$ for validation.
The ADAM solver \cite{ADAM} was chosen for training, due to its automatic learning rate adjustment. 
As suggested in \cite{ADAM}, we chose the exponential decay rates to $\beta_1=0.9$ and $\beta_2=0.999$. 
The base learning rate was chosen to $l=0.0001$.
The training process was stopped after about $1.6$ epochs ($100000$ iterations).

Input space reduction to gain computation speed and save memory is very common in machine learning approaches with image data. 
However, simply down scaling the covariance matrices in \DOGMA\ cells is formally wrong, due to the assumption of independent cell objects \cite{NUSS2016} in the fusion approach.
Furthermore, velocity differences in neighboring cells represent a local velocity variance, which can be exploited by the convolutional network architecture.
This makes the single cell covariance redundant, which is stored in the last three channels in $\GMchannels$.
Therefore, we chose to reduce the input space by omitting the velocity covariance matrices and use only the channels $\MassOcc$, $\MassFree$, $v_\East$, and $v_\North$.
The prediction target times were chosen to $\SI{0.5}{s},\SI{1}{s},\SI{1.5}{s},...,\SI{3}{s}$.
%

% *** Section 7
\section{Evaluation}
\label{sec:results}

The network was trained and tested in a downtown scenario with numerous dynamic objects.
Due to the shared space situation where pedestrians and bikes cross the road at arbitrary locations, the prediction task is particularly challenging.
The location also confronts us with occlusions in the field of view.
The experimental vehicle provides \SI{360}{\degree} perception using an IBEO LUX long-range laser scanner covering about \SI{200}{m} in the front with an opening angle of about \SI{100}{\degree} and four Velodyne VLP-16 laser scanners, each providing $16$ layers laser measurement ranging up to \SI{100}{m}.
The \DOGMA\ covers $901\times901$ cells with a cell width of $\SI{0.15}{m}$.
To evaluate the prediction, we used about $\SI{12}{min}$ from $4$ sequences. 
Videos showing the prediction results can be watched under \href{https://youtu.be/iw-eXnnjfj0}{https://youtu.be/iw-eXnnjfj0}.

% % % % % % % % % % % % % % % % % % % % % % % % % % % % % % %
%
% Comparison to particle filter
%
% % % % % % % % % % % % % % % % % % % % % % % % % % % % % %
%
We compare the learning based prediction to a particle based algorithm.
The occupancy probability and velocity distribution in $\GMchannels$ of a grid cell $\gridcell$ is represented by particles with state vector $\mathbf{x}=\left[E, N, v_\East, v_\North \right]^\mathrm{T}$. 
We calculate a long-term prediction by forward propagating the particles with the constant velocity model.
Only particles of non-static cells are forward propagated. 
Cells are considered non-static if the velocity variance is below $\SI{3}{m^2/s^2}$ and the velocity magnitude is greater than $\SI{0.7}{m/s}$. 
These parameters turned out to yield good separation between static and dynamic regions, however, their heuristic nature is a drawback compared to the learning based approach.
In our experiments, $900000$ particles were sampled and forward propagated at dynamic cells to provide a dense distribution of predicted occupancy.
%

% % % % % % % % % % % % % % % % % % % % % % % % % % % % % % % %
% ROC-Curve
% % % % % % % % % % % % % % % % % % % % % % % % % % % % % % % % 

\begin{figure}
\centering
\vspace{0.8mm}
\setlength\figureheight{3.5cm}
\setlength\figurewidth{7.5cm}
\input{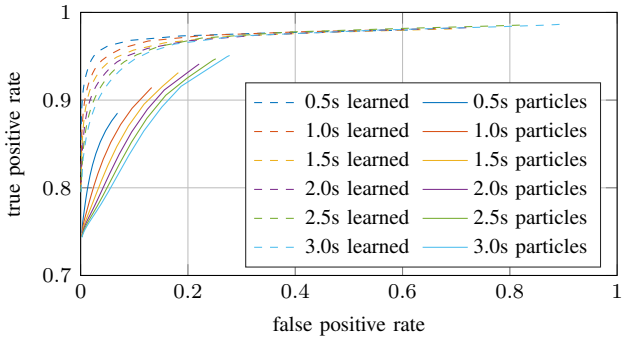}
\caption{
Receiver operating characteristic curves for different prediction times of the learning and particle based approach.
}
\label{fig:roc}
\end{figure}

The overall prediction performance is illustrated in Fig.~\ref{fig:roc} showing a receiver operating characteristic (ROC) curve for the learned and particle approach with different prediction times. 
For this plot, each dynamic cell is classified as occupied or free with the varying threshold parameter $\rocthreshold\in(0,1)$.
It is important to note, that the two approaches are not compared against automatically generated labels, but to the actually perceived \DOGMA\ at the corresponding prediction time including static and dynamic cells.
To calculate true positive and false negative rates in Fig.~\ref{fig:roc}, the perceived \DOGMA\ is thresholded with $P_{\Occupied}>0.55$, using a fixed value slightly over the ambiguous value of $0.5$.
Also, static parts in the prediction approaches were thresholded with this fixed value, while the varying threshold is only applied to dynamic prediction.
The dynamic neural network channels are directly thresholded with $\rocthreshold$, while for the particle approach the prediction is considered occupied if the number of particles per cell $n_{\gridcell}$ is greater than $\rocthreshold \cdot \max(n_{\gridcell})$.
%

% % % % % % % % % % % % % % % % % % % % % % % % % % % % % % % %
% ROC-Curve PLOT
% % % % % % % % % % % % % % % % % % % % % % % % % % % % % % % % 

In general, it can be observed, that for an increasing prediction time, a growing false positive rate (fpr) has to be taken into account to reach a common true positive rate (tpr), which means that the uncertainty of the predicted area grows with increasing prediction time.
The curve start points at $\mathrm{fpr}=0$ illustrate the case when dynamic prediction is completely ignored. 
At this point it can be seen, that in the particle approach, a minimum tpr of $0.75$ can already be reached by just predicting static cells. 
The learning based approach, however, reaches a minimum tpr of $0.81$, indicating better performance for classifying static regions in the grid map.

The top right ends of the curves illustrate the ROC for $\rocthreshold=0$, while the cases $n_\gridcell=0$ (no particle in a cell) and $P_\Occupied=0$ are not considered occupied due to the grater condition.
The limitation of the tpr for both approaches means, that false negatives cannot be eliminated. Examples are given below.  
However, the maximum tpr at a prediction time of $\SI{0.5}{s}$ lies at $98\%$ for the learning based approach and $89\%$ for the particle approach.
A longer prediction time causes higher uncertainty which leads to a growing area covered by particles or predicted occupied by the network. 
This in turn leads to an increasing true positive rate at the expense of a growing false positive rate.
Thus, the maximum true positive rate for the particle filter lies at $95\%$ at $\SI{3}{s}$ prediction horizon and all particle based ROC curves never reach the worst ROC curve of the learning based approach.

\subsection{Predicting Multi-Modal Scene Evolution}

% % % % % % % % % % % % % % % % % % % % % % % % % % % % % % % % %
% Multi Modal Prediction Image
% % % % % % % % % % % % % % % % % % % % % % % % % % % % % % % % %
% % % % % % % % % % % % % % % % % % % % % % % % % % % % % % % % %
% Multi Modal
% % % % % % % % % % % % % % % % % % % % % % % % % % % % % % % % %
\begin{figure*}
\vspace{1.4mm}
\captionsetup[subfigure]{font=footnotesize,labelformat=empty,labelfont=scriptsize}
\centering
\begin{subfigure}[t]{\textwidth}
	\centering
	\includegraphics[trim=0 0 0 0,clip,height=3cm]{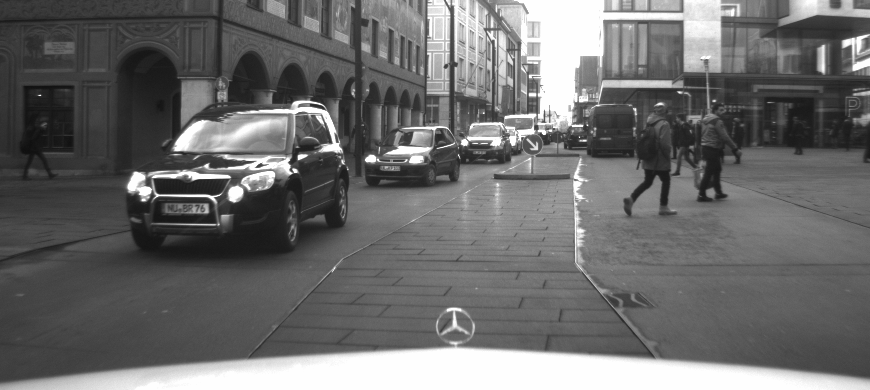}
	\begin{tikzpicture}[overlay]
		\draw [dash pattern={on 7pt off 3pt}, line width = 0.5mm, red] (-3.6,1.8) ellipse (0.6 and 0.5);
	\end{tikzpicture}
	\caption{}
\end{subfigure}

\def\shroistartX{354}
\def\shroistartY{266}
\def\shroiWidth{141}
\def\shroiHeight{189}
\def\shifigurewidth{0.135}
\begin{subfigure}[b]{\shifigurewidth\textwidth}
	\includegraphics[width=\textwidth]{img/scenario4/DLR_001_001_11065.tikz}
	\caption{input}
\end{subfigure}
\begin{subfigure}[b]{\shifigurewidth\textwidth}
	\includegraphics[width=\textwidth]{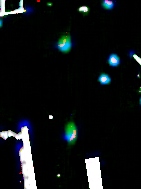}
	\caption{$\SI{0.5}{s}$}
\end{subfigure}
\begin{subfigure}[b]{\shifigurewidth\textwidth}
		\includegraphics[width=\textwidth]{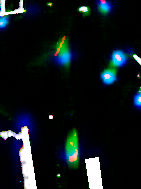}
	\caption{$\SI{1.0}{s}$}
\end{subfigure}
\begin{subfigure}[b]{\shifigurewidth\textwidth}
		\includegraphics[width=\textwidth]{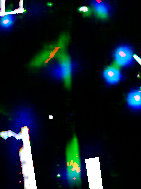}
	\caption{$\SI{1.5}{s}$}
\end{subfigure}
\begin{subfigure}[b]{\shifigurewidth\textwidth}
		\includegraphics[width=\textwidth]{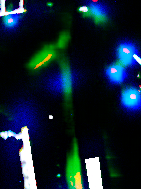}
	\caption{$\SI{2.0}{s}$}
\end{subfigure}
\begin{subfigure}[b]{\shifigurewidth\textwidth}
		\includegraphics[width=\textwidth]{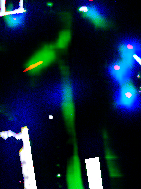}
	\caption{$\SI{2.5}{s}$}
\end{subfigure}
\begin{subfigure}[b]{\shifigurewidth\textwidth}
		\includegraphics[width=\textwidth]{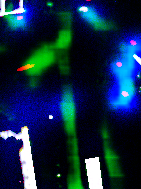}
	\caption{$\SI{3.0}{s}$}
\end{subfigure}
\caption{Prediction results for a (possibly) turning vehicle (dashed, red ellipse) before turning maneuver.
The excerpt of the input grid map (bottom left) shows the cell velocity estimates with orange lines, illustrating that it is not clear at input time, if the vehicle will turn or go straight.
Predicted occupancy from \SI{0.5}{s} to \SI{3.0}{s} is illustrated in RGB images in the bottom row.
True occupancy is encoded in the red channel, learned prediction in the green, and prediction with particles in the blue channel. Overlapping results are encoded as mixed RGB color, e.g., white in static regions.
Starting from \SI{1.0}{s}, the multi-modality of the predicted spatial occupancy distribution is visible in the learned prediction result, while the particles only cover the straight maneuver.
Also, the static edges of the wall, misclassified as dynamic, are predicted as moving by the particle approach but not by the learning approach.
}
\label{fig:Result1_multimodal_static}
\end{figure*}

Fig. \ref{fig:Result1_multimodal_static} shows an example of long-term prediction results in a complex scenario.
More specifically, it depicts the image of the on-board camera, showing parallel traffic with a possible right-hand turn for oncoming traffic. 
The bottom left image in Fig. \ref{fig:Result1_multimodal_static} shows the vehicle surrounding represented by a \DOGMA.
The cell colors and overlaid orange lines visualize the filtered velocity of moving cells.
The displayed \DOGMA\ is used as only input to the prediction network.
The following RGB images in the bottom row visualize the prediction result at times \SI{0.5}{s} to \SI{3.0}{s}.
The color channels in these images encode the different approaches.
The true occupancy is encoded in the red channel, the output of the neural net appears in the green channel, and the particle prediction is illustrated in the blue channel. 
An overlap of the prediction leads to a mixed RGB-color, ideally white when both approaches overlap with true occupancy, as for example at most static regions.
For better visualization, static cells are only drawn if $P_\Occupied>0.6$.

Note, that at the given time instance, it is not obvious whether the vehicle marked with a red ellipse will be taking a right-hand turn or pass the junction.
Hence, the prediction network outputs a bimodal occupancy distribution (green) for the vehicle's trajectory, while propagated particles (blue) only cover the straight maneuver, based on the initial velocity distribution. 
This can be seen in Fig. \ref{fig:Result1_multimodal_static} within the prediction horizon \SI{1.0}{s} to \SI{3.0}{s}. 
The RGB color appears yellow if only the learned prediction overlaps with the true occupancy.
These results indicate that the prediction network is capable of inferring possible driving maneuvers and even the spatio-temporal occupancy distribution of future trajectories. % without explicit knowledge of the road topology.

\subsection{Static Regions} 
\begin{figure*}
\vspace{1.4mm}
\captionsetup[subfigure]{labelformat=empty, font=footnotesize,labelfont=scriptsize}
\centering
\begin{subfigure}[t]{\textwidth} %
	\centering
	\includegraphics[trim=0 0 0 0,clip,height=3cm]{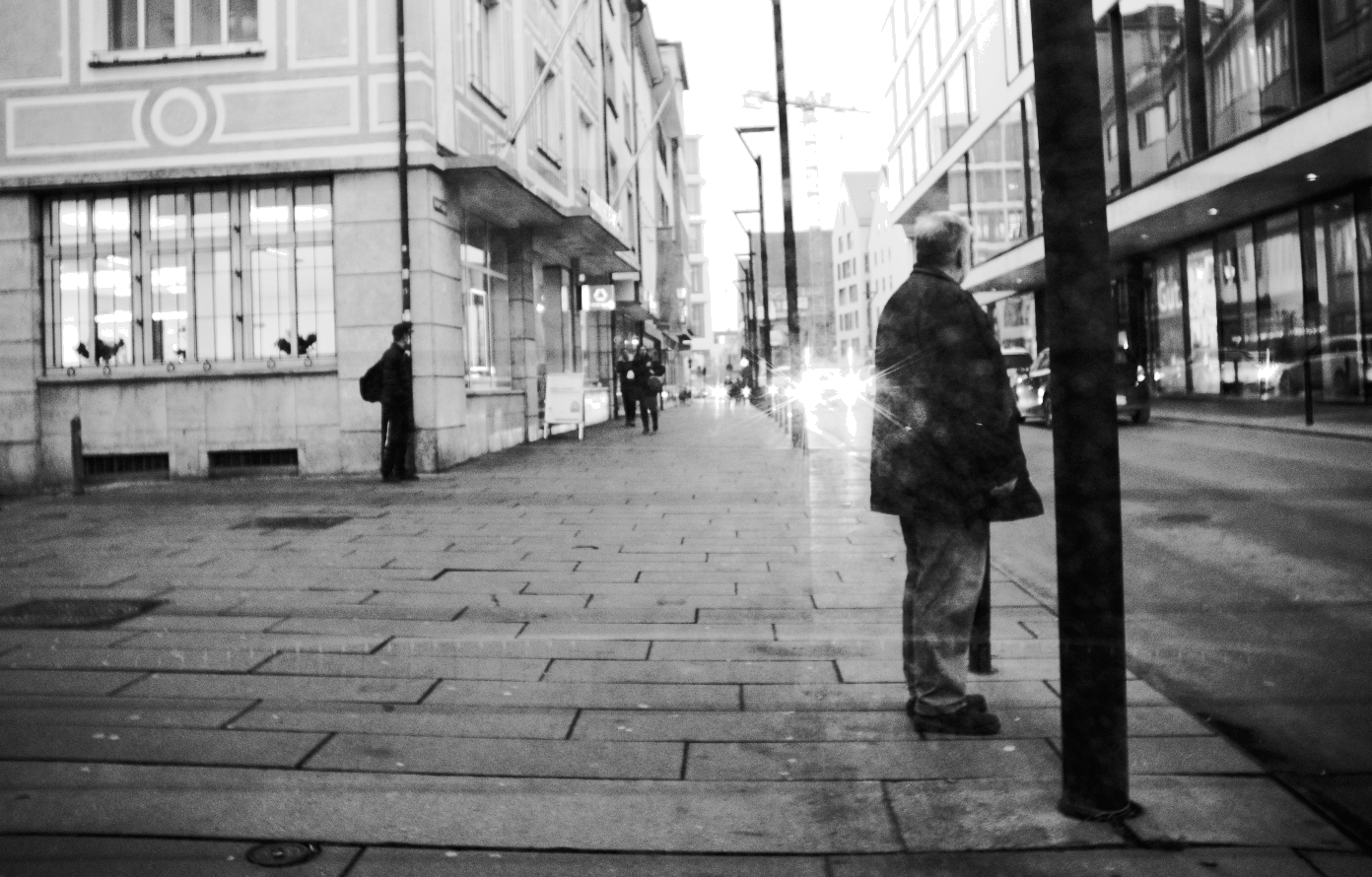}
	\includegraphics[trim=0 0 0 0,clip,height=3cm]{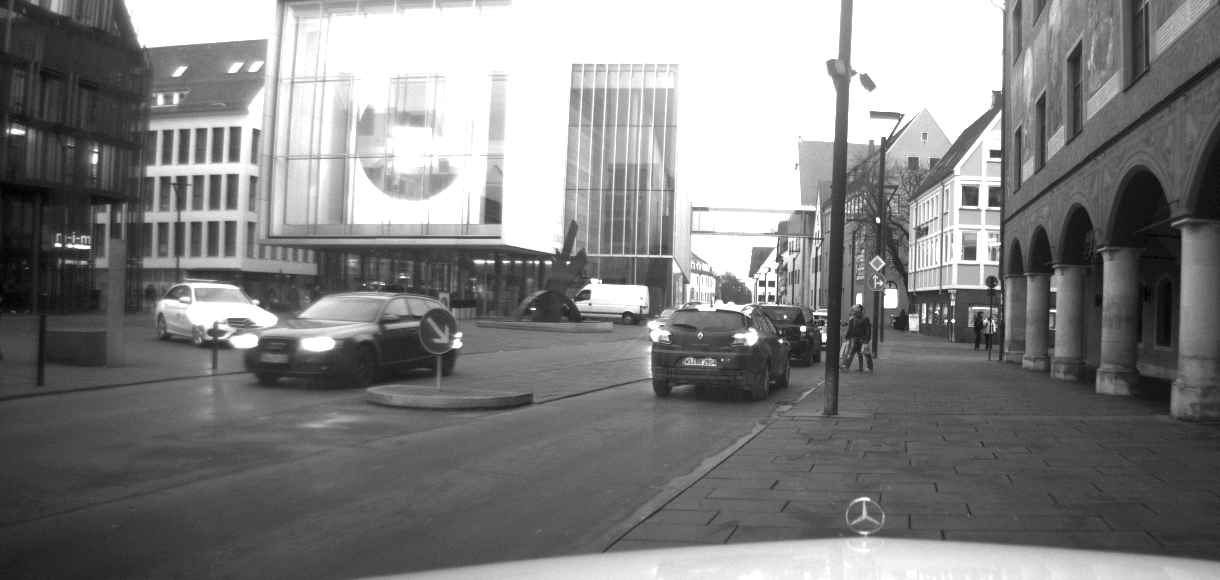}
	\begin{tikzpicture}[overlay]
		\draw [dash pattern={on 7pt off 3pt on 2pt off 3pt}, line width = 0.5mm, blue] (-7.875,1.35) ellipse (0.6 and 1);
		\draw [dash pattern={on 7pt off 3pt}, line width = 0.5mm, red] (-4.65,1.275) ellipse (1 and 0.4);
	\end{tikzpicture}
	\caption{}
\end{subfigure}

\def\shroistartX{421}
\def\shroistartY{370}
\def\shroiWidth{107}
\def\shroiHeight{204}
\def\shifigurewidth{0.135}
\def\shvdist{1pt}
\begin{subfigure}[b]{\shifigurewidth\textwidth}
	\includegraphics[width=\textwidth]{img/scenario8/pred_input.tikz}
	\vspace{\shvdist}
	
	\includegraphics[width=\textwidth]{img/scenario8/pred_input_2.tikz}
	\caption{input}
\end{subfigure}
\vspace{1em}
\begin{subfigure}[b]{\shifigurewidth\textwidth}
	\includegraphics[width=\textwidth]{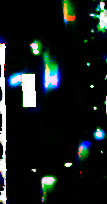}
	\vspace{\shvdist}
	
	\includegraphics[width=\textwidth]{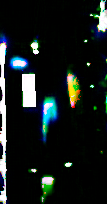}
	\caption{$\SI{0.5}{s}$}
\end{subfigure}
\begin{subfigure}[b]{\shifigurewidth\textwidth}
	\includegraphics[width=\textwidth]{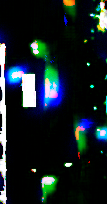}
	\vspace{\shvdist}
	
	\includegraphics[width=\textwidth]{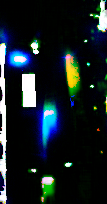}
	\caption{$\SI{1.0}{s}$}
\end{subfigure}
\begin{subfigure}[b]{\shifigurewidth\textwidth}
	\includegraphics[width=\textwidth]{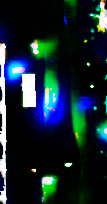}
	\vspace{\shvdist}
	
	\includegraphics[width=\textwidth]{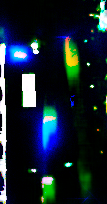}
	\caption{$\SI{1.5}{s}$}
\end{subfigure}
\begin{subfigure}[b]{\shifigurewidth\textwidth}
	\includegraphics[width=\textwidth]{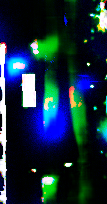}
	\vspace{\shvdist}
	
	\includegraphics[width=\textwidth]{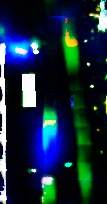}
	\caption{$\SI{2.0}{s}$}
\end{subfigure}
\begin{subfigure}[b]{\shifigurewidth\textwidth}
	\includegraphics[width=\textwidth]{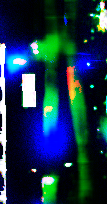}
	\vspace{\shvdist}
	
	\includegraphics[width=\textwidth]{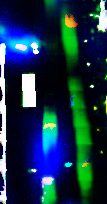}
	\caption{$\SI{2.5}{s}$}
\end{subfigure}
\begin{subfigure}[b]{\shifigurewidth\textwidth}
	\includegraphics[width=\textwidth]{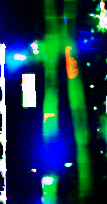}
	\vspace{\shvdist}
	
	\includegraphics[width=\textwidth]{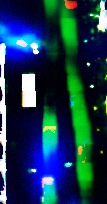}
	\caption{$\SI{3.0}{s}$}
\end{subfigure}
\caption{Prediction result for a possibly crossing pedestrian (blue, dash-dotted ellipses). 
The first row shows the ego vehicle's rear and front camera images, the other rows show the input \DOGMA\ and prediction result with the same color encoding as in Fig.~\ref{fig:Result1_multimodal_static}. 
In the 2nd row, the network (green channel) predicts a high possibility of the pedestrian to cross the street, while the approaching upwards driving vehicle (red, dashed ellipses) would arrive in front of the pedestrian after about $\SI{3}{s}$ (last column).
$\SI{1.8}{s}$ later in the 3rd row, the approaching vehicle is correctly predicted to arrive already in $\SI{1}{s}$ (3rd column) and the predicted crossing path of the pedestrian vanished.
The particle filter (blue channel) predicts no movement of the pedestrian due to it's static classification, and only parts of the vehicle in the 3rd row, due to partially occlusion. 
The prediction results also show a vehicle moving down approaching a standing vehicle. The network is able to predict a stopping maneuver, while the particle approach predicts a collision.
}
\label{fig:pedestrian_not_crossing}
\end{figure*}

Distinguishing between static and dynamic cells is a hard task considering only cell-wise features. 
The particle prediction, as well as the automatic label generation algorithm contain misclassifications of static cells, especially at the border of static regions.
In the particle based prediction, misclassification due to bad parameters leads either to blurred occupancy prediction of static areas, or to missing predictions of dynamic objects. 
An example can be seen in Fig. \ref{fig:Result1_multimodal_static}.
The corners of the static wall are surrounded by predicted particles, drawn in blue.
The neural network in contrast, can deal with misclassified cells in the automatic labeling process by simply not predicting the movement of static regions.
However, the static waiting pedestrian in Fig. \ref{fig:pedestrian_not_crossing} is predicted to possibly cross the street by the learning based approach, while the particle approach can only predict its movement as soon as it moves.

% % % % % % % % % % % % % % % % % % % % % % % % % % % % % % % % %
% Pedestrian not crossing
% % % % % % % % % % % % % % % % % % % % % % % % % % % % % % % % %
% % % % % % % % % % % % % % % % % % % % % % % % % % % % % % % % %
% Richtung Neu-Ulm
% % % % % % % % % % % % % % % % % % % % % % % % % % % % % % % % %

\subsection{Predicting Partially Occluded Objects}
Another pattern, typical for dynamic occupancy grids, is that for objects entering the field of view, the area estimated as occupied tends to grow.
From initially only a few occupied cells, a solid silhouette can be seen when the object is deeper in the field of view and the filter converges. 
The same applies for vanishing occlusion.
The neural network learned to predict an enlarged occupied area, even when only a few pixels are initially occupied and dynamic in the \DOGMA. 
The last row in Fig. \ref{fig:pedestrian_not_crossing} shows a vehicle driving upwards in the \DOGMA\ input image which is partially occluded by a downwards driving vehicle.
The prediction result of the neural network (green), covers the entire vehicle, while the particle approach (blue) only predicts the vehicle front, underestimating the future occupied area.
Object detection could improve the particle approach, considering object size and occlusions.
However, object detection can be challenging when objects just enter the field of view \cite{steyer2017object}.

\subsection{Reducing Prediction Variance with Object Interactions}
% % % % % % % % % % % % % % % % % % % % % % % % % % % % % % % % %
% More Interaction
% % % % % % % % % % % % % % % % % % % % % % % % % % % % % % % % %
% % % % % % % % % % % % % % % % % % % % % % % % % % % % % % % % %
% inception
% % % % % % % % % % % % % % % % % % % % % % % % % % % % % % % % %
\begin{figure}[h!]
\vspace{1.5mm}
\captionsetup[subfigure]{labelformat=empty, font=footnotesize,labelfont=scriptsize}
\centering

\def\shroistartX{378}
\def\shroistartY{306}
\def\shroiWidth{132}
\def\shroiHeight{144}
\def\shifigurewidth{0.43}
\def\shvdist{2mm}

\begin{subfigure}[b]{\shifigurewidth\linewidth}
	\includegraphics[width=\textwidth]{img/inception/pred_input.tikz}
	\caption{original input}
\end{subfigure}
\begin{subfigure}[b]{\shifigurewidth\linewidth}
\centering
	\includegraphics[width=\textwidth]{img/inception/pred_input_2.tikz}
	\caption{manipulated input}
\end{subfigure}
\vspace{1em}

% % % % %
% 0.5 seconds
% % % % %
\if false 
\begin{subfigure}[b]{\linewidth}
\centering
	\includegraphics[width=\shifigurewidth\linewidth]{img/inception/pred_2.tikz}
	\includegraphics[width=\shifigurewidth\linewidth]{img/inception/pred_2_2.tikz}
	\caption{$\SI{0.5}{s}$}
\end{subfigure}
%\vspace{1em}
%
\fi

\begin{subfigure}[b]{\linewidth}
\centering
	\includegraphics[width=\shifigurewidth\linewidth]{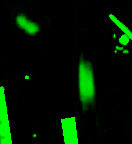}
    \includegraphics[width=\shifigurewidth\linewidth]{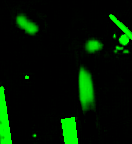}
	\begin{tikzpicture}[overlay]
		%\draw [red] (0,0) grid (-15,4);
		\draw [red] (-7.67,2.35) to (-0.12, 2.35);
	\end{tikzpicture}
	\caption{$\SI{1.0}{s}$}
\end{subfigure}

\vspace{\shvdist}

\begin{subfigure}[b]{\linewidth}
\centering
	\includegraphics[width=\shifigurewidth\linewidth]{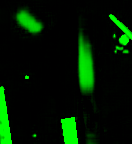}
	\includegraphics[width=\shifigurewidth\linewidth]{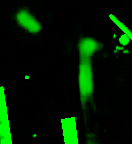}
	\begin{tikzpicture}[overlay]
		\draw [red] (-7.67,2.55) to (-0.12, 2.55);
	\end{tikzpicture}
	\caption{$\SI{1.5}{s}$}
\end{subfigure}

\vspace{\shvdist}

\begin{subfigure}[b]{\linewidth}
\centering
	\includegraphics[width=\shifigurewidth\linewidth]{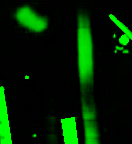}
	\includegraphics[width=\shifigurewidth\linewidth]{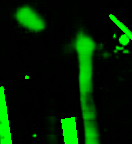}
	\begin{tikzpicture}[overlay]
		\draw [red] (-7.67,3.05) to (-0.12, 3.05);
	\end{tikzpicture}
	\caption{$\SI{2.0}{s}$}
\end{subfigure}

\if false
\vspace{\shvdist}

\begin{subfigure}[b]{\linewidth}
\centering
	\includegraphics[width=\shifigurewidth\linewidth]{img/inception/pred_6.tikz}
	\includegraphics[width=\shifigurewidth\linewidth]{img/inception/pred_6_2.tikz}
	\begin{tikzpicture}[overlay]
		\draw [red] (-8.02,3.45) to (-0.12, 3.45);
	\end{tikzpicture}
	\caption{$\SI{2.5}{s}$}
\end{subfigure}
\fi

\if false

\vspace{\shvdist}

\begin{subfigure}[b]{\linewidth}
\centering
	\includegraphics[width=\shifigurewidth\linewidth]{img/inception/pred_7.tikz}
	\includegraphics[width=\shifigurewidth\linewidth]{img/inception/pred_7_2.tikz}
	\caption{$\SI{3.0}{s}$}
\end{subfigure}
\fi
\caption{Prediction result with (left column) and without interaction (right column). The right column shows the same \DOGMA{} input, but with a pedestrian (top row; blue, dash-dotted ellipse) artificially added from another sequence. The presence of the pedestrian influences the predicted occupancy of the approaching upwards driving vehicle. Red horizontal lines were added for improved visual reference.
}
\label{fig:inception2}
\end{figure}
A main motivation for the use of a CNN is the ability to exploit spatial context, an advantage to model object interactions.
Interactions reduce the future trajectory space, e.g., due to social forces \cite{Luber2010} or traffic rules.
An example can be observed in the car following scenario illustrated in Fig. \ref{fig:pedestrian_not_crossing}.
The figure includes two moving vehicles: 
the left one in the input \DOGMA, appearing purple, drives downwards approaching a standing car.
The right one, appearing light green in the input \DOGMA{} and marked with a red dashed ellipse, drives upwards on a free lane.
Prediction on the free lane is very uncertain after $\SI{2}{s}$, resulting in a large area predicted to possibly be occupied.
On the blocked lane, in contrast, the network predicts the vehicle to slow down, reducing the future scene variance and causing a smaller region to be predicted as occupied.
The particle filter overestimates the vehicle's motion, resulting in a collision with the standing object.

The scene also illustrates an example of pedestrian-vehicle interactions.
A pedestrian stands close to the road, his intention is not clear, and the vehicle marked with the red ellipse is approaching from his right-hand side. 
The last two rows illustrate the scene prediction at two different starting times. 
In the first prediction, the approaching vehicle is still about $\SI{3}{s}$ away from the pedestrian, and a possible crossing trajectory is predicted by the neural network.
Starting the prediction $\SI{1.8}{s}$ later (bottom row), the pedestrian still didn't move, but the approaching vehicle came closer.
For the pedestrian it would be hard to cross the road now, since the vehicle arrives at the pedestrian in about $\SI{1}{s}$, true and predicted.
Appropriately, the neural network prediction for the pedestrian adapted and no crossing trajectory is predicted anymore.
The particle approach, in contrast, cannot predict any movement for temporarily static objects.

The ability of the trained CNN to model interactions is further investigated in the following experiment.
Rather than using only real-world data with a single observable outcome, manipulating the input can yield useful information on the predictive power of the CNN.
Therefore, we added objects to a given scene in order to force the model to react on the new scenario.
Fig.~\ref{fig:inception2} shows side by side the prediction of two almost identical scenes of an upwards driving vehicle.
In the right column, however, a pedestrian was artificially added off-road approaching the lane.
The blue dash-dotted ellipse highlights the input modification.
The insertion was accomplished by extracting the \DOGMA{} representation of a pedestrian from another recording and placing it into the scene of interest.
Horizontal reference lines (red) illustrate the difference in occupancy prediction at points of interest.
The predictions of the upwards driving car in the original scenario do not show any explicit interactions with objects in its environment, and is consequently used as reference occupancy to study the role of interactions.
By examining the predictions at $\SI{1.0}{s}$, in case of the augmented input, the car already does not advance as much as in respect to the original case.
This predicted deceleration of the vehicle, can only be a result of interaction with the added pedestrian.
Inspecting the further prediction results, the effect becomes even more obvious. 
At $\SI{1.5}{s}$, the original prediction of the vehicle already exceeds the spatial position of the pedestrian included in the other case, which would lead to a collision if interaction is not modeled.
But on the contrary, the network reacts to the presence of the pedestrian by a slowed down predicted movement, which can be perceived by the less likely occupancy in the area between the car and the pedestrian ($\SI{1.5}{s}$).
Most obviously, the vehicle will not run over the pedestrian, causing no occupancy behind it, while, in the case of non-existence, the car is predicted to just go through ($\SI{2.0}{s}$).
In conclusion, in the augmented scenario, the predicted occupancy caused by the vehicle alters, which can only be due to the particular presence of the velocities and occupancies of the pedestrian.
%

% *** Section 8 - conclusion ***
\section{Conclusions}
\label{sec:conclusion}

We presented a learning based situation prediction approach predicting the \SI{360}{\degree} perceivable scene in a single neural network.
Beside prediction, the network performs a segmentation of static and dynamic areas.
The machine learning approach benefits from a fully automatic label generation method, yielding an unsupervised character.
We introduced a loss function that counteracts imbalanced pixels of different categories and trained the dataset with rare dynamic cells.
The prediction was compared to a particle based approach.
The results showed, that the network can predict highly complex scenarios with various road users of different class up to \SI{3}{s}.
It can be seen, that the network is able to consider different maneuver classes, e.g., turn right or go straight, and interactions between road users reduce the prediction uncertainty.
In ongoing studies, the loss function is extended to consider the probabilistic nature of prediction by adjusting the pixel-wise weighting factors according to increasing variance of an object's future position.
Experiments with other network architectures, i.e., recurrent neural networks are planned.
%

% *** acknowledgements ***
\section*{Acknowledgements}

The research leading to these results has received funding from the European Union under the H2020 EU.2.1.1.7.\ ECSEL Programme, as part of the RobustSENSE project, contract number 661933.
%
%Responsibility for the information and views set out in this publication lies entirely with the authors. 
Responsibility for the information and views in this publication lies entirely with the authors. 
The authors would like to thank all RobustSENSE partners. % for their cooperation and valuable contribution.

%%%%%%%%%%%%%%%%%%%%%%%%%%%%%%%%%%%%%%%%%%%%%%%%%%%%%%%%%%%%%%%%%%%%%%%%%%%%%%%%
\bibliographystyle{IEEEtran}
\bibliography{IEEEtranControl,SH.bib} 

\begin{thebibliography}{10}
\providecommand{\url}[1]{#1}
\csname url@samestyle\endcsname
\providecommand{\newblock}{\relax}
\providecommand{\bibinfo}[2]{#2}
\providecommand{\BIBentrySTDinterwordspacing}{\spaceskip=0pt\relax}
\providecommand{\BIBentryALTinterwordstretchfactor}{4}
\providecommand{\BIBentryALTinterwordspacing}{\spaceskip=\fontdimen2\font plus
\BIBentryALTinterwordstretchfactor\fontdimen3\font minus
  \fontdimen4\font\relax}
\providecommand{\BIBforeignlanguage}[2]{{%
\expandafter\ifx\csname l@#1\endcsname\relax
\typeout{** WARNING: IEEEtran.bst: No hyphenation pattern has been}%
\typeout{** loaded for the language `#1'. Using the pattern for}%
\typeout{** the default language instead.}%
\else
\language=\csname l@#1\endcsname
\fi
#2}}
\providecommand{\BIBdecl}{\relax}
\BIBdecl

\bibitem{kesting2008car_following_models}
A.~Kesting and M.~Treiber, ``Calibrating car-following models by using
  trajectory data: Methodological study,'' \emph{Transportation Research
  Record: Journal of the Transportation Research Board}, no. 2088, pp.
  148--156, 2008.

\bibitem{petrich2013mapbasedprediction}
D.~Petrich \emph{et~al.}, ``Map-based long term motion prediction for vehicles
  in traffic environments,'' in \emph{IEEE Conference on Intelligent
  Transportation Systems}, 2013, pp. 2166--2172.

\bibitem{hoermann2017}
S.~Hoermann, D.~Stumper, and K.~Dietmayer, ``Probabilistic long-term prediction
  for autonomous vehicles,'' in \emph{IEEE Intelligent Vehicles Symposium},
  June 2017, pp. 237--243.

\bibitem{odhams2004speedincurves}
A.~M. Odhams and D.~J. Cole, ``Models of driver speed choice in curves,'' in
  \emph{Proceedings of the 7th International Symposium on Advanced Vehicle
  Control}, 2004.

\bibitem{Graf_Intersection_2014}
R.~Graf \emph{et~al.}, ``A learning concept for behavior prediction at
  intersections,'' in \emph{IEEE Intelligent Vehicles Symposium}, June 2014,
  pp. 939--945.

\bibitem{Kuhnt2016}
F.~Kuhnt \emph{et~al.}, ``Understanding interactions between traffic
  participants based on learned behaviors,'' in \emph{IEEE Intelligent Vehicles
  Symposium}, June 2016, pp. 1271--1278.

\bibitem{wiest2017PHD}
J.~Wiest, ``Statistical long-term motion prediction,'' Ph.D. dissertation, Ulm
  University, 2017.

\bibitem{Laengkvist201411}
M.~Laengkvist, L.~Karlsson, and A.~Loutfi, ``A review of unsupervised feature
  learning and deep learning for time-series modeling,'' \emph{Pattern
  Recognition Letters}, vol.~42, pp. 11 -- 24, 2014.

\bibitem{barshalom2001OutOfSequence}
Y.~Bar-Shalom, ``Update with out-of-sequence measurements in tracking: exact
  solution,'' \emph{IEEE Transactions on Aerospace and Electronic Systems},
  vol.~38, no.~3, pp. 769--777, Jul 2002.

\bibitem{aeberhard2012AsynchronousSensors}
M.~Aeberhard \emph{et~al.}, ``Track-to-track fusion with asynchronous sensors
  using information matrix fusion for surround environment perception,''
  \emph{IEEE Transactions on Intelligent Transportation Systems}, vol.~13,
  no.~4, pp. 1717--1726, Dec 2012.

\bibitem{agrawal2013stockpredictiontechniques}
J.~Agrawal, V.~Chourasia, and A.~Mittra, ``State-of-the-art in stock prediction
  techniques,'' \emph{International Journal of Advanced Research in Electrical,
  Electronics and Instrumentation Engineering}, vol.~2, no.~4, pp. 1360--1366,
  2013.

\bibitem{tsai2010stockprediction}
C.-F. Tsai and Y.-C. Hsiao, ``Combining multiple feature selection methods for
  stock prediction: Union, intersection, and multi-intersection approaches,''
  \emph{Decision Support Systems}, vol.~50, no.~1, pp. 258--269, 2010.

\bibitem{fama1965stockprediction_isguessing}
E.~F. Fama, ``The behavior of stock-market prices,'' \emph{The journal of
  Business}, vol.~38, no.~1, pp. 34--105, 1965.

\bibitem{bollen2011twitter_predicts_stock}
J.~Bollen, H.~Mao, and X.~Zeng, ``Twitter mood predicts the stock market,''
  \emph{Journal of computational science}, vol.~2, no.~1, pp. 1--8, 2011.

\bibitem{gruhl2005predictstockbyonlinechat}
D.~Gruhl \emph{et~al.}, ``The predictive power of online chatter,'' in
  \emph{ACM SIGKDD international conference on Knowledge Discovery in Data
  Mining}, 2005, pp. 78--87.

\bibitem{NUSS2016}
D.~Nuss \emph{et~al.}, ``A random finite set approach for dynamic occupancy
  grid maps with real-time application,'' \emph{arXiv preprint
  arXiv:1605.02406}, 2016.

\bibitem{Elfes1989}
A.~Elfes, ``Using occupancy grids for mobile robot perception and navigation,''
  \emph{Computer}, vol.~22, no.~6, pp. 46--57, 6 1989.

\bibitem{Thrun2005}
S.~Thrun, W.~Burgard, and D.~Fox, \emph{Probabilistic Robotics (Intelligent
  Robotics and Autonomous Agents)}.\hskip 1em plus 0.5em minus 0.4em\relax The
  MIT Press, 2005.

\bibitem{steyer2017object}
S.~Steyer, G.~Tanzmeister, and D.~Wollherr, ``Object tracking based on
  evidential dynamic occupancy grids in urban environments,'' in \emph{IEEE
  Intelligent Vehicles Symposium}.\hskip 1em plus 0.5em minus 0.4em\relax IEEE,
  2017, pp. 1064--1070.

\bibitem{dempster2008generalization}
A.~P. Dempster, ``A generalization of bayesian inference,'' in \emph{Classic
  works of the dempster-shafer theory of belief functions}.\hskip 1em plus
  0.5em minus 0.4em\relax Springer, 2008, pp. 73--104.

\bibitem{noh2015deconvnet}
H.~Noh, S.~Hong, and B.~Han, ``Learning deconvolution network for semantic
  segmentation,'' in \emph{2015 IEEE International Conference on Computer
  Vision (ICCV)}, Dec 2015, pp. 1520--1528.

\bibitem{shelhamer2017FCNs}
E.~Shelhamer, J.~Long, and T.~Darrell, ``Fully convolutional networks for
  semantic segmentation,'' \emph{IEEE transactions on pattern analysis and
  machine intelligence}, vol.~39, no.~4, pp. 640--651, 2017.

\bibitem{long2015FCN}
J.~Long, E.~Shelhamer, and T.~Darrell, ``Fully convolutional networks for
  semantic segmentation,'' in \emph{IEEE Conference on Computer Vision and
  Pattern Recognition}, 2015, pp. 3431--3440.

\bibitem{zeiler2014FCNvis}
M.~D. Zeiler and R.~Fergus, ``Visualizing and understanding convolutional
  networks,'' in \emph{European conference on computer vision}.\hskip 1em plus
  0.5em minus 0.4em\relax Springer, 2014, pp. 818--833.

\bibitem{badrinarayanan2015segnet}
V.~Badrinarayanan, A.~Kendall, and R.~Cipolla, ``Segnet: A deep convolutional
  encoder-decoder architecture for scene segmentation,'' \emph{IEEE
  Transactions on Pattern Analysis and Machine Intelligence}, vol.~PP, no.~99,
  pp. 1--1, 2017.

\bibitem{zeiler2011adadeconv}
M.~D. Zeiler, G.~W. Taylor, and R.~Fergus, ``Adaptive deconvolutional networks
  for mid and high level feature learning,'' in \emph{IEEE International
  Conference on Computer Vision}, 2011, pp. 2018--2025.

\bibitem{ADAM}
D.~Kingma and J.~Ba, ``Adam: A method for stochastic optimization,''
  \emph{arXiv preprint arXiv:1412.6980}, 2014.

\bibitem{Luber2010}
M.~Luber \emph{et~al.}, ``People tracking with human motion predictions from
  social forces,'' in \emph{IEEE International Conference on Robotics and
  Automation}, May 2010, pp. 464--469.

\end{thebibliography}

\end{document}